\documentclass[journal,twoside,web]{IEEETran}
\usepackage{cite}
\usepackage{amsmath,amssymb,amsfonts}
\usepackage{algorithmic}
\usepackage{graphicx}
\usepackage{textcomp}
\usepackage{amsmath,amssymb,amsfonts}
\usepackage{array}
\usepackage{graphicx}
\usepackage{textcomp}
\usepackage{subcaption}
\usepackage{bm}
\usepackage{soul}
\usepackage{booktabs}
\usepackage{tikz} 
\usepackage[colorlinks=true, allcolors=blue]{hyperref}
\usepackage{bbding}
\usepackage{pifont}
\usepackage{paralist}
\usepackage{comment}
\usepackage{color}

\newcommand{\be}{\begin{equation}}
\newcommand{\ee}{\end{equation}}
\newcommand{\bn}{\begin{numerate}}
\newcommand{\en}{\end{numerate}}
\newcommand{\bi}{\begin{itemize}}
\newcommand{\ei}{\end{itemize}}
\newcommand{\bmt}{\begin{bmatrix}}
\newcommand{\emt}{\end{bmatrix}}
\def\BibTeX{{\rm B\kern-.05em{\sc i\kern-.025em b}\kern-.08em
    T\kern-.1667em\lower.7ex\hbox{E}\kern-.125emX}}
\begin{document}
\title{MR.CAP: Multi-robot Joint Control and Planning for Object Transport}
\author{Hussein Ali Jaafar, Cheng-Hao Kao, and Sajad Saeedi
\thanks{Toronto Metropolitan University, 
 {husseinali.jaafar@torontomu.ca}}}

\maketitle
\thispagestyle{empty}
\begin{abstract}
With the recent influx in demand for multi-robot systems throughout industry and academia, there is an increasing need for faster, robust, and generalizable path planning algorithms. Similarly, given the inherent connection between control algorithms and multi-robot path planners, there is in turn an increased demand for fast, efficient, and robust controllers. We propose a scalable joint path planning and control algorithm for multi-robot systems with constrained behaviours based on factor graph optimization. We demonstrate our algorithm on a series of hardware and simulated experiments. Our algorithm is consistently able to recover from disturbances and avoid obstacles while outperforming state-of-the-art methods in optimization time, path deviation, and inter-robot errors. See the code and supplementary video for experiments\footnote{\href{https://h2jaafar.github.io/projects/mrcap/}{https://h2jaafar.github.io/projects/mrcap/}}.
\end{abstract}


\section{Introduction}

\IEEEPARstart{I}n recent years, there has been a growing demand for cohesive real-time multi-robot systems across various industries, including agriculture, warehousing, and hospitality~\cite{Parker2009}. These industries have increasingly relied on robotic systems to enhance workflow efficiency and cost-effectiveness. 
Key challenges in these {multi-robot} systems revolve around problems such as path planning, control, and estimation. 
Both planning and control are inherently forward-looking tasks, making use of {the} current state to predict information on future states. This similarity drives an attempt to integrate these two into one, {which leads to} reduce{d} inter-module communication, resulting in faster response times and holistically optimal results~\cite{STEAP}.

In multi-robot systems, the integration of 
path planning and control into a cohesive real-time framework presents significant challenges. These difficulties manifest as communication overheads, computational inefficiencies, and scalability limits {with respect to the number of obstacles and robots}. 
{Problem formulation and optimization is the key factor distinguishing the existing solutions~\cite{halsted2021survey}.
These solutions vary depending on the problem specifics and the required quality. For collaborative object transport~\cite{mclurkin2015}, hierarchical quadratic programming
(HQP)~\cite{hal-geometric-approach} performs partial joint planning and control, where the planning is based only on the current view of the sensor, unlike our work, which takes into account the full map of the environment. Further, scalability and handling uncertainties remain open problems.}

To address these challenges, this paper introduces an innovative scalable end-to-end joint planning and control framework. Our approach seamlessly integrates planning and control into a unified optimization framework, harnessing the versatility of factor graphs~{\cite{factor-graphs-for-perception}}. 
The utilization of factor graphs, with its probabilistic structure, equips our system to handle a wide spectrum of scenarios.
Notably, our approach prioritizes scalability, making it easy to accommodate larger robot teams {and complex environments} without introducing substantial computational complexity. 
We rigorously evaluate our algorithm against current state-of-the-art approaches, including {model predictive control} (MPC)~{\cite{AUGLAG, Powell1994ADS}} and HQP{~\cite{hal-geometric-approach}, with} simulated and hardware-based experiments. {The formulation and experiments are designed for collaborative object transport; however, it is flexibly extendable to a wide range of problems.} 
{See the project page for demos, a 3D example, and the code.}

\begin{figure}[t!]
    \centering
     \includegraphics[width=\linewidth]{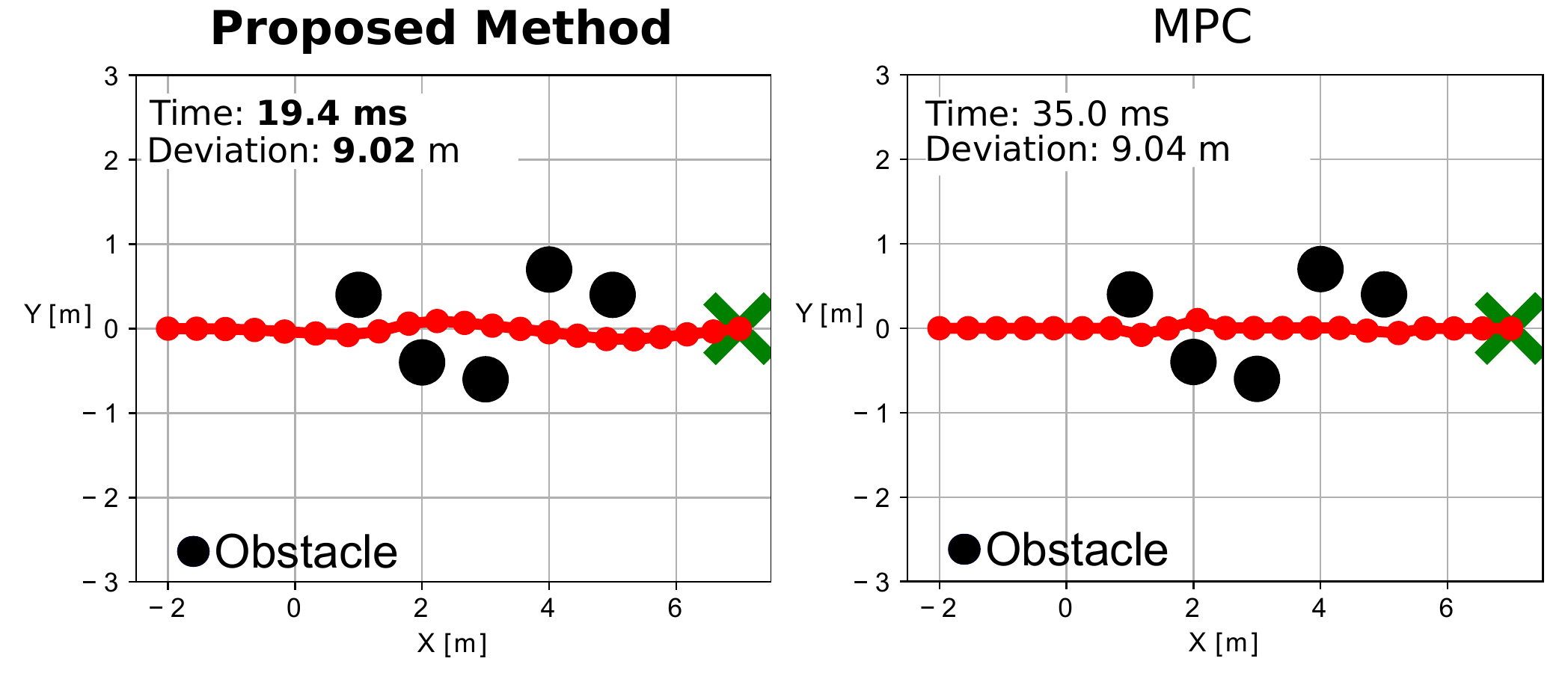}
    \begin{minipage}{.5\columnwidth}
    \centering
    \includegraphics[width=0.8\linewidth]{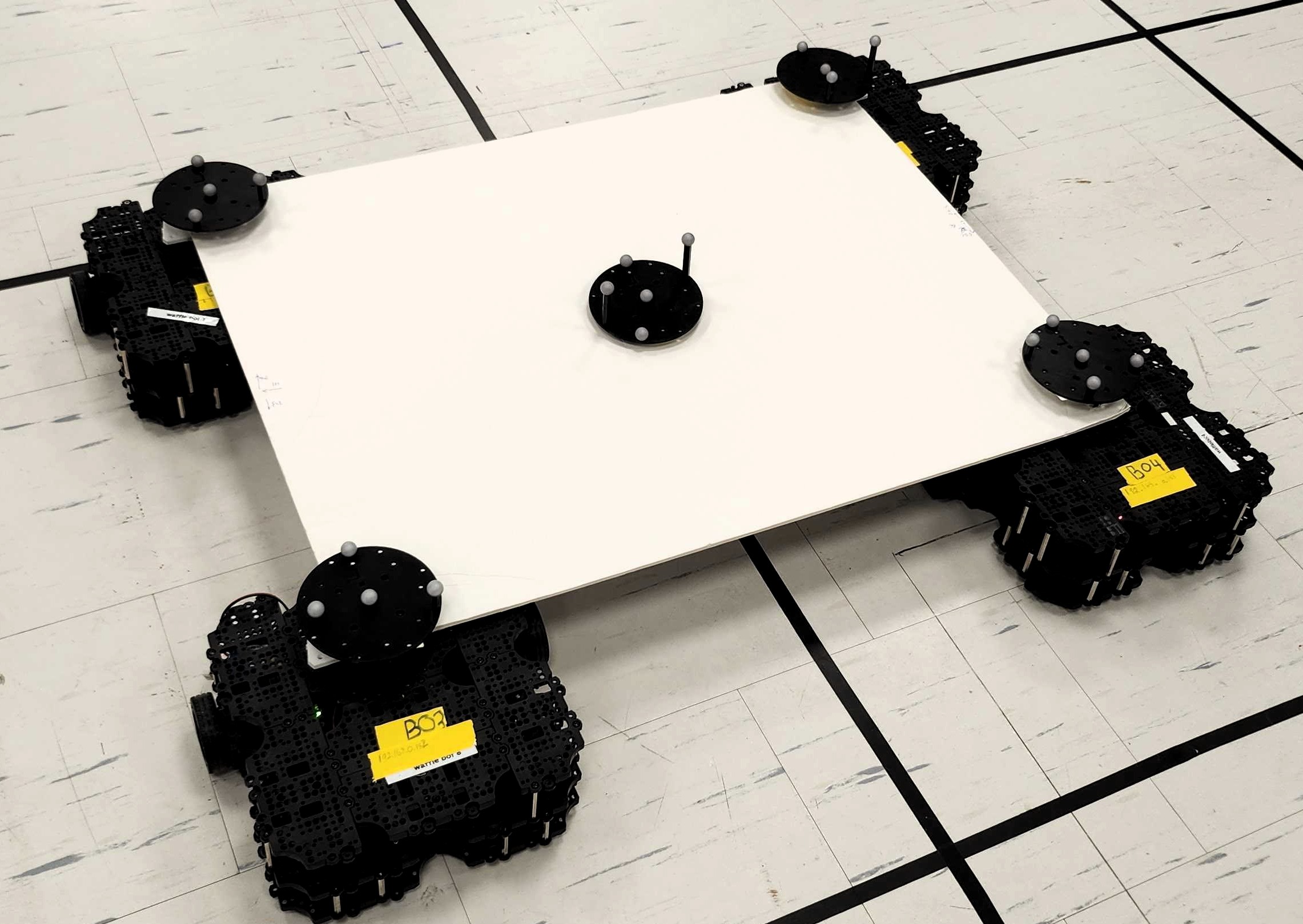}
    \end{minipage}%
    \begin{minipage}{0.5\columnwidth}
     \caption{Four robots moving a payload in a complex environment. Our method outperforms model predictive control, MPC (constraint-based) in time and path length.} \label{fig:two-robots}
    \end{minipage}
\end{figure}

We focus on joint control and planning for collaborative object transport. The initial formation is assumed to be known. Fig.~\ref{fig:two-robots} shows a demo where four robots \textcolor{black}{transport an object while avoiding obstacles.}
\textcolor{black}{Our method is faster than the MPC-based methods~\cite{AUGLAG} and generates a shorter path.}
We present the literature review in Sec.~\ref{sec:lit_rev}. Sec.~\ref{sec:method} presents our method, followed by experiments in Sec.~\ref{sec:experiments}, and conclusions in Sec.~\ref{sec:conclusions}.

\section{Literature Review}\label{sec:lit_rev}

Motion planning is a cornerstone of robotic systems. 
Planners can be categorized into several types: 
graph-based, e.g. A*~\cite{astar_og} and Dijkstra~\cite{choset2005principles};
sampling-based, e.g. RRTs~\cite{rrt} and PRM~\cite{kavraki1994probabilistic}; and  
trajectory optimizers, e.g. covariant Hamiltonian optimization for motion planning (CHOMP)~\cite{chomp}, T-CHOMP~\cite{T-CHOMP}, Multigrid CHOMP~\cite{multi-chomp}, STOMP~\cite{STOMP}. {Optimizers are typically probabilistic and use \emph{maximum likelihood}, as in \cite{toussaint2009}, or \emph{maximum a posteriori} (MAP), as in simultaneous trajectory estimation and planning (STEAP)~\cite{STEAP}.}
{Some of these methods have high discretization costs, which can be overcome by methods in~\cite{gpmp2, TrajOpt2}}. 
{Additionally, to reduce the complexity of the methods, and} present trajectories with fewer states, Gaussian process motion planning (GPMP)~\cite{gpmp} represents the continuous-time trajectory as a sample from a Gaussian process. 
In many recent planning algorithms, factor graphs play a key role, e.g. \cite{bazzana_handling_2022} and \cite{DFG}. {Planning in dynamic settings (see ITOMP~\cite{ITOMP}) and lack of guarantee for optimality and feasibility of paths remain an active research field.}

In real-world applications, the collaboration of multiple robots {is inevitable}. 
There are several research frontiers in such systems, including 
cooperative exploration {\cite{patwardhan2023distributed}}, 
localization \cite{murai2022robotweb, spasojevic_active_nodate}, 
formation {\cite{guerrero-bonilla_formations_2017}}, 
planning \cite{gnn, cao_path_nodate, gbp-planner}, and
control~\cite{pierson_cooperative_2016}. 
A widely used controller is model predictive control (MPC)~\cite{dellaert-fg-approach}. {An active research area in these frontiers is to solve the problems in a distributed/decentralized manner, as done in \cite{saravanos_distributed_nodate} for control, or in \cite{yan_decentralized_2021} for manipulation.}

An extension to the multi-robot control and planning problem is object {transport/pushing/}manipulation~\cite{HE20209859}, using multiple robots to transport an object such that it undergoes a desired motion or tracks a desired path \cite{ogasawara1996}. 
Various approaches are used for cooperative object transport. 
Examples include 
furniture moving algorithm \cite{jennings1995}, 
using virtual structures \cite{lujak_distributed_2010}, 
combining primitive controllers \cite{mclurkin2015}, 
hybrid controller with a receding horizon planner \cite{daniela2015}, 
constrained optimization  \cite{hal-geometric-approach}, and 
various other interesting centralized and decentralized algorithms \cite{schwager2015}, \cite{schwager2016}, \cite{schwagerants2016},  \cite{culbertson_decentralized_2018}. 
{Data-driven models making use of behavioral systems theory have also been proposed for multi-robot quadrupedal locomotion~\cite{data-distributed},~\cite{data-driven-ral}. Distributed scalable manipulation and planning inspired by separable optimization variable ADMM (SOVA) methods have been investigated in ~\cite{schwager2020-control-planning},~\cite{schwager2021trajopt}. An open problem is the capacity to handle uncertainties, essential for real-world applications.}  

\section{{Proposed} Method} \label{sec:method}
We tackle a joint planning and control problem of a multi-robot for {object transport (see Fig.~\ref{fig:sys})}. The robots collaboratively manipulate the {object} centroid to keep it along a planned path and avoid obstacles in an online fashion. For obstacle avoidance, we require that the entire payload avoid the obstacles along with the robots. The formation of the robots is assumed to be known and remains fixed.

\subsection{{Motion Model}}

The task of joint planning and control is discretized into $N$ steps, i.e. waypoints. There are {a total of $J$} obstacles to be avoided. 
There are two types of variables: the states $\mathbf{x}=\mathbf{x_{0:N}}$, $\mathbf{x}_n = \{x_n, y_n, \theta_n\}$, $n=1...N$ {(see Fig.~\ref{fig:sys})} and controls $\mathbf{u}=\mathbf{u_{0:N-1}}$, $\mathbf{u}_n = {\{v_n, \omega_n\}}$. 
Control efforts and velocities in the global frame $\mathbf{v}=\mathbf{v_{0:N-1}}$, $\mathbf{v}_n = \{\dot{x}_n, \dot{y}_n, \dot{\theta}_n\}$ can easily be related by using kinematic Jacobians. 

We use a number of $I$ differential drive robots, which can be arranged in an arbitrary pattern underneath the payload. 
Let the superscripts $c$ and $r_i$ represent the centroid and the $i$-th robot in the formation respectively, we have centroid poses $\mathbf{x}^c$, centroid {controls $\mathbf{u}^c$}, robot poses $\mathbf{x}^{r_i}$, and robot {controls $\mathbf{u}^{r_i}$}. {Assuming small integration intervals and constant velocity within each step, the motion model of a robot is obtained by integration via the Runge-Kutta method \cite{bazzana_handling_2022}}

\be
\label{eq: robot motion model}
    \mathbf{x}_{n + 1}^{r_i} =
    \mathbf{x}_{n}^{r_i} + T_s
    \bmt
        cos(\theta_{n}^{r_i} + \frac{\omega_{n}^{r_i}}{2}) & 0 \\
        sin(\theta_{n}^{r_i} + \frac{\omega_{n}^{r_i}}{2}) & 0 \\
        0 & 1
    \emt
    \mathbf{u}_n^{r_i}, 
\ee
\vspace{-0.5 cm}
\begin{align*}
    \forall i={1,2,\dots,I} \quad\text{and}\quad
    \forall n={1,2,\dots,N}~.
\end{align*}
{$T_s$ is the processing time step. The motion model of the centroid takes the same form as Eq.~\eqref{eq: robot motion model}. However,}  due to the nonholonomic constraints imposed by the differential drive robots used and the geometric constraints embedded in the system design, we split the motion of the centroid into two stages; translation and rotation. {This eliminates the need for Runge-Kutta approximations and improves the path shape of the system. As such, the motion model of the centroid \textcolor{black}{is}:

\be
\label{eq: simplified robot motion model}
    \mathbf{x}_{n + 1}^{c} =
    \mathbf{x}_{n}^{c} + T_s
    \bmt
        cos(\theta_{n}^{c}) & 0 \\
        sin(\theta_{n}^{c}) & 0 \\
        0 & 1
    \emt
    \mathbf{u}_n^{c}~.
\ee

\begin{figure}[t]
   \centering
    \begin{minipage}{.5\columnwidth}
    \centering
    \includegraphics[width=0.75\linewidth]{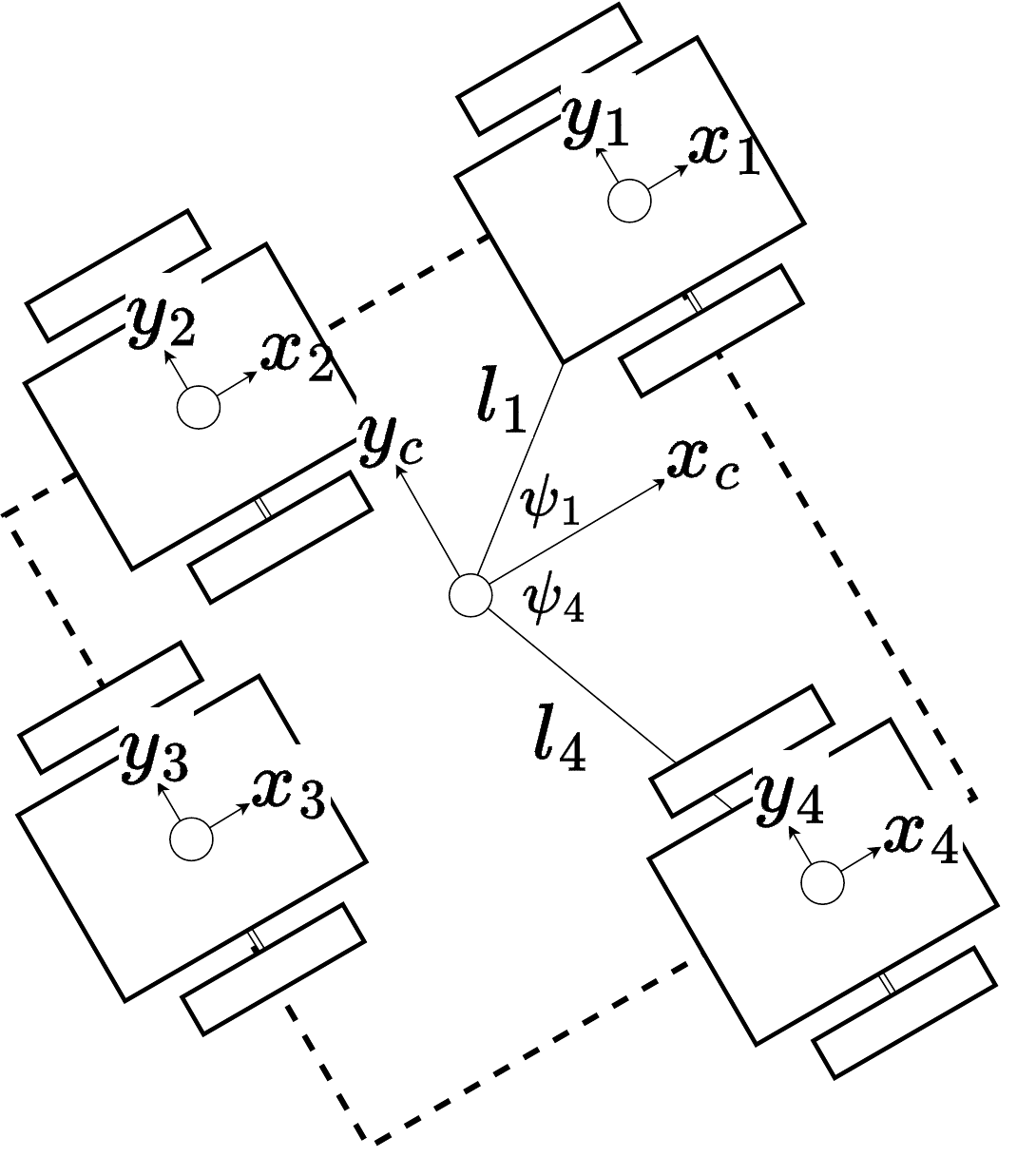}
    \vspace{-0.9 cm}
    \end{minipage}%
    \begin{minipage}{0.5\columnwidth}
    \caption {Example of four robots arranged to move the payload placed at an arbitrary location on the platform. The system is capable of operating with more robots, \textcolor{black}{for} heavy payloads. {The robots can have} any starting orientation when connected, as the optimizer accounts for any needed adjustments. 
    }
        \label{fig:sys}
\vspace{-.9 cm}
    \end{minipage}
\end{figure}

{Since there are no actuators on the payload, one could also optimize for the centroid velocity $\mathbf{v}_n^{c}$ in the world frame to further simplify the problem. This means:
}
\be
    \mathbf{x}_{n + 1}^{c} =
    \mathbf{x}_{n}^{c} + T_s
    \mathbf{v}_n^{c}~.
\ee

{Once the control of the centroid is optimized, the robots pivot until they reach their required orientations $\theta_n^{r_i}$ and then apply the controls $\mathbf{u}_n^{r_i}$:}

\begin{equation}
\label {eq: cases}
\theta_n^{r_i} = 
\begin{cases}
  \theta_n^{c}, &\text{if centroid translates} \\
  \theta_n^{c} + \psi_i \pm \pi/2, &\text{if centroid rotates}
\end{cases}~,
\end{equation}

\begin{equation}
    \label{eq: relating motion models}
    \mathbf{u}_n^{r_i} 
    = 
    \bmt
        1 & l_i \\
        0 & 1
    \emt
    \mathbf{u}_n^{c}~.
\end{equation}

{As discussed, the motion of the system is split into pure rotation and pure translation. The required orientations of the robots $\theta_n^{r_i}$ and robot controls $\mathbf{u}_n^{r_i}$ change depending on the centroid's motion. The sign in Eq.~\eqref{eq: cases} depends on the direction of the centroid's rotation. The first and second columns of the matrix in Eq.~\eqref{eq: relating motion models} correspond to the centroid's translation and rotation, respectively. Depending on the specific formation and system dynamics of a system, Eq.~\eqref{eq: relating motion models} can be modified and extended. The conversion between the centroid velocity $\mathbf{v}_n^{c}$ and centroid control $\mathbf{u}_n^{c}$ is trivial and omitted for conciseness.} 
{These reductions allow for efficient optimization while improving scalability dramatically. It also reduces the inter-robot communication while still able to achieve impressive performance, as shown in experiments.}

\subsection{Factor Graphs}
We use factor graphs to perform efficient and unified optimization of the joint control and planning problem. Factor graphs are a bipartite graphical structure that represents a factorized function, such that variable relationships are encapsulated by edges and factors. We represent our path optimization problem as a MAP inference problem, and use GTSAM~\cite{gtsam}, to perform inference over the factor graph. 

We can represent the centroid poses, $\mathbf{x}^c$, and the centroid velocity $\mathbf{u}^c$ as variables in the factor graph. We develop relationships between these variables and use sparse optimizers to solve for our control efforts. We found that Levenberg–Marquardt (LM) is a solver well-suited for most of our use cases, as it provides optimization results of decent accuracy and great robustness. {LM solver provides fast and local convergence guarantees~\cite{LM_OG_PAPER}, which are important to real-time systems.}
{As shown in Fig.~\ref{fig:multi-robot-fg}, factors used are:} {$f_n^x$ unary pose factor, penalizes the system for deviating away from} {the} {provided {initial} path; $f_n^u$ unary speed factor, constrains the speed at which the system moves along the planned path to the motor speed limit; $f_n^{m}$ ternary motion factor, forces the algorithm to adhere to the established motion model; and $f_n^{obs_j}$ obstacle avoidance factor, penalizes the system for a path nearing obstacles.} The overall cost function is
\begin{align}
    J(&\mathbf{x}, \mathbf{u}) = \sum_{n = k}^N e_n^x + \sum_{n = k}^{N - 1}( e_n^u +  e_n^{m}) + {\sum_{j = 1}^{J}\sum_{n = {k + 1}}^{N} e_n^{obs_j}}~, \\
    &e_n^x(\mathbf{x}_{n}^c) = {||\mathbf{x}_{n}^c - \mathbf{x}_{n}^{c, ref}||}_{\Omega_n^x}^2~, \\
    &e_n^u(\mathbf{u}_n^c) = {||\mathbf{u}_n^c - {\mathbf{u}_{n}^{c, ref}}||}_{\Omega_n^u}^2~, \\
    &e_n^{m}(\mathbf{x}_n^c, \mathbf{u}_n^c, {\mathbf{x}_{n+1}^c}) = {||{\mathbf{x}_{n + 1}^c}(\mathbf{x}_n^c, \mathbf{u}_n^c)^* - \mathbf{x}_{n + 1}^c||}_{\Omega_n^{m}}^2~, \\
    &e_n^{obs_{{j}}}(\mathbf{x}_n^c) = 
    \begin{cases}
      {||1 - \frac{d_j}{R}||}_{\Omega_n^{obs}}^2, &d_j < R \\
      {0}, &d_j \geq R
    \end{cases}~.
\end{align}

{$||.||$  is the Mahalanobis distance, with $\Omega$ terms representing the information matrices of the variables.} $e_n^x$ and $e_n^u$ are the costs corresponding to factors $f_n^x$ and $f_n^u$, respectively. $e_n^{m}$ is the cost corresponding to factors $f_n^{m}$, where ${\mathbf{x}_{n + 1}^c}(\mathbf{x}_n^c, \mathbf{u}_n^c)^*$ represents the estimated pose ${\mathbf{x}_{n + 1}^c}^*$ to which the centroid moves over the next time step provided the optimized pose $\mathbf{x}_n^c$ and control $\mathbf{u}_n^c$. $e_n^{obs}$ is the defined cost corresponding to factors $f_n^{obs}$, {where $d_j$ represents the centroid's distance from obstacle $j$ and $R$ represents the radius of a predefined safety bubble that encapsulates the entire multi-robot system~\cite{gbp-planner}. Factors can be added, removed, or modified depending on the task requirements}. Subtractions {are} overloaded by appropriate manifold operations handled in GTSAM.

{With proper selection of the covariances}, the equivalent effects of regulation matrix adjustments can be achieved {and terms that are commonly considered constraints (motion model and obstacle) can be well approximated with soft constraints as have been shown in similar factor graph implementations in~\cite{gbp-planner} and~\cite{bazzana_handling_2022}. It was also proven in ~\cite{Yang_2021} that covariances of 0 yield deterministic solutions that satisfy the boundaries confined by the constraints. There are other alternatives utilizing barrier functions~\cite{barrierMPC}, Lagrange multipliers, and dual graph optimization~\cite{Xie2020AFA}, but we find the formulation presented above sufficient for our application.}


To {analyze the complexity}, we perform a worst-case complexity analysis on our approach and MPC. For our approach, we create a factor graph with $O(2n + 4n)$ complexity, 2 variables, and 4 factors are added at each step, $n$. The LM solver with multifrontal QR decomposition can be approximated $ \Tilde{O}(\epsilon^{-2})$~\cite{bergou_convergence_2020} where $\Tilde{O}(.)$ indicates the presence of logarithmic factors in $\epsilon$. Post-optimization extractions take $O(I)$, as such, $\Tilde{O}(n\epsilon^{-2}(n^2+m) + nI)$ where $m$ are solver parameters. We can develop the MPC formulation in a similar manner, with the complexity of Augmented Lagrangian solver taken as $O(\left|\log(\epsilon)\right|^2 \epsilon^{-2})$ ~\cite{auglag_complexity}, the complexity becomes $O(nh({\mid \! \log ( \epsilon )\!\mid}^2 \epsilon ^{-2}) + nI)$, where $h$ are the horizon steps. Both methods demonstrate polynomial scaling with the number of steps and an inverse-square relationship with error tolerance. However, our emphasis on scalability concerns the number of robots and environmental complexity rather than just task steps, N. In these aspects, our approach exhibits a linear behavior, and yields shorter optimization times compared to MPC, effectively enhancing performance for larger numbers of robots and more complex environments.

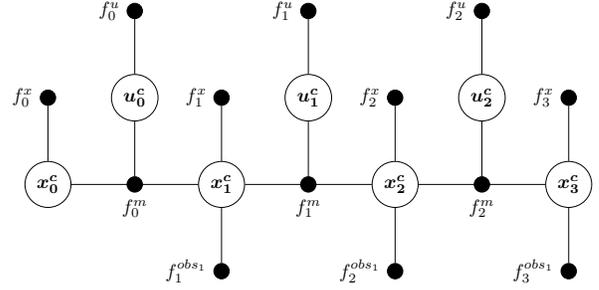
\begin{figure}[t]
\centering
\resizebox{0.9\linewidth}{!}{
\begin{tikzpicture}[node distance = {16mm}, main/.style = {draw, circle}] 
\tikzstyle{point1}=[circle, draw=black,fill=black, inner sep=0.1cm]
\tikzstyle{motor_up}=[circle, draw=blue,fill=blue, inner sep=0.1cm]
\tikzstyle{vicon}=[circle, draw=green,fill=green, inner sep=0.1cm]

\node[main] (x10) {$\bm{x_{0}^c}$}; 
\node[point1] (f0) [right of =x10, label=below:$f_0^{m}$]{};
\node[main] (x11) [right of=f0]{$\bm{x_{1}^c}$}; 
\node[point1] (f1) [right of =x11, label=below:$f_1^{m}$]{};
\node[main] (x12) [right of=f1] {$\bm{x_{2}^c}$};
\node[point1] (f2) [right of =x12, label=below:$f_2^{m}$]{};
\node[main] (x13) [right of=f2] {$\bm{x_{3}^c}$};

\node[main] (u0) [above of =f0] {$\bm{u_{0}^c}$}; 
\node[main] (u1) [above of =f1] {$\bm{u_{1}^c}$}; 
\node[main] (u2) [above of =f2] {$\bm{u_{2}^c}$}; 

\node[point1] (fu0) [above of =u0, label=left:$f_0^u$]{};
\node[point1] (fu1) [above of =u1, label=left:$f_1^u$]{};
\node[point1] (fu2) [above of =u2, label=left:$f_2^u$]{};

\node[point1] (fx0) [above of =x10, label = left:$f_0^x$]{};
\node[point1] (fx1) [above of =x11, label = left:$f_1^x$]{};
\node[point1] (fx2) [above of =x12, label = left:$f_2^x$]{};
\node[point1] (fx3) [above of =x13, label = left:$f_3^x$]{};

\draw (x10) -- (x11); \draw (x11) -- (x12); \draw (x12) -- (x13);
\draw (u0) -- (f0); \draw (u1) -- (f1); \draw (u2) -- (f2); 
\draw (u0) -- (fu0); \draw (u1) -- (fu1); \draw (u2) -- (fu2); 

\node[point1] (fo1) [below of =x11, label=left:$f_{1}^{obs_1}$]{};
\node[point1] (fo2) [below of =x12, label=left:$f_{2}^{obs_1}$]{};
\node[point1] (fo3) [below of =x13, label=left:$f_{3}^{obs_1}$]{};

\draw (x11) -- (fo1); \draw (x12) -- (fo2); 
\draw (x10) -- (fx0); \draw (x11) -- (fx1); \draw (x12) -- (fx2); \draw (x13) -- (fx3); \draw (x13) -- (fo3); 

\end{tikzpicture} 
}
    \caption{Multi-robot control and planning factor graph with centroid pose variables and centroid control variables. Factors and their associated costs are defined in the text. Since estimation is not the emphasis of this work, only the variables at the current state and future states are used in the optimization problem for each iteration. A finite horizon is used in our case, extending to the terminal state to ensure long-term stability.}
    \label{fig:multi-robot-fg}
\end{figure}

\section{Experiments}\label{sec:experiments}

{Experiments are performed in three environments:} i) a pure-simulated environment, ii) a physics-based Gazebo simulation~\cite{gazebo}, and iii) real-world hardware. 
The performances are evaluated by {five} metrics: 
1) average deviation, 
2) inter-robot errors, 
3) optimization time, 
4) path length, and 
5) distance to goal. 
The average deviation indicates the average distance between each state and its corresponding state {on the \emph{initial path} estimate}. 
{The \emph{initial path} is a straight-line path from the start to the goal point. It will be optimized eventually.}
{We assume that initial robot formation is performed beforehand, and as such,} the inter-robot error describes the amount by which distances between robots change when no payloads are present. {This measures the impact of Eq.~\eqref{eq: relating motion models} on the system.} 
The optimization time measures the time taken to optimize the paths and obtain individual robot control efforts. {All experiments were run on an Intel i7-10700 at 4.8 GHz, with 16 GB DDR4 memory.} The path length is a metric for assessing the system's ability to reach the end state following the shortest path, even when perturbed or blocked by an obstacle. {The distance to goal metric {evaluates} the system's ability to reach the goal and {ensures} short {paths} do not get over-prioritized.}

\subsection{Experiment 1: Simulation}

Simulations {in Experiment 1} are designed to test the scalability by increasing the number of robots, {from 4 to 128, and the complexity of the environment by introducing more obstacles.} It is implemented {in} C++, with metrics including optimization time, path deviation, and path length. 

Here the baseline is model predictive control (MPC), based on NLopt~\cite{NLopt}. We implement two MPC algorithms; gradient and gradient-free approaches: augmented Lagrangian AUGLAG~\cite{AUGLAG} and constrained optimization by linear approximation COBYLA~\cite{Powell1994ADS}. The MPC was designed to handle system requirements as either constraints or penalties (penalties are implemented similarly to our approach), where the former provides more accurate results and the latter optimizes faster. We define these as MPC-C and MPC-P, respectively. As {shown} in Table~\ref{fig:equality}, both approaches are used in our comparative analysis. The best parameters for all algorithms were chosen by performing an exhaustive search over all the parameters within 2000 iterations (see supplementary), such that the distance to the goal is below 0.06$m$ and the optimization time does not exceed 80$ms$. In total, 20580 parameter sets were examined. See Table~\ref{fig: parameters} for the best parameters.

\begin{figure}[b!]
    \centering
    \begin{subfigure}[b]{0.49\linewidth}
        \includegraphics[width=\linewidth]{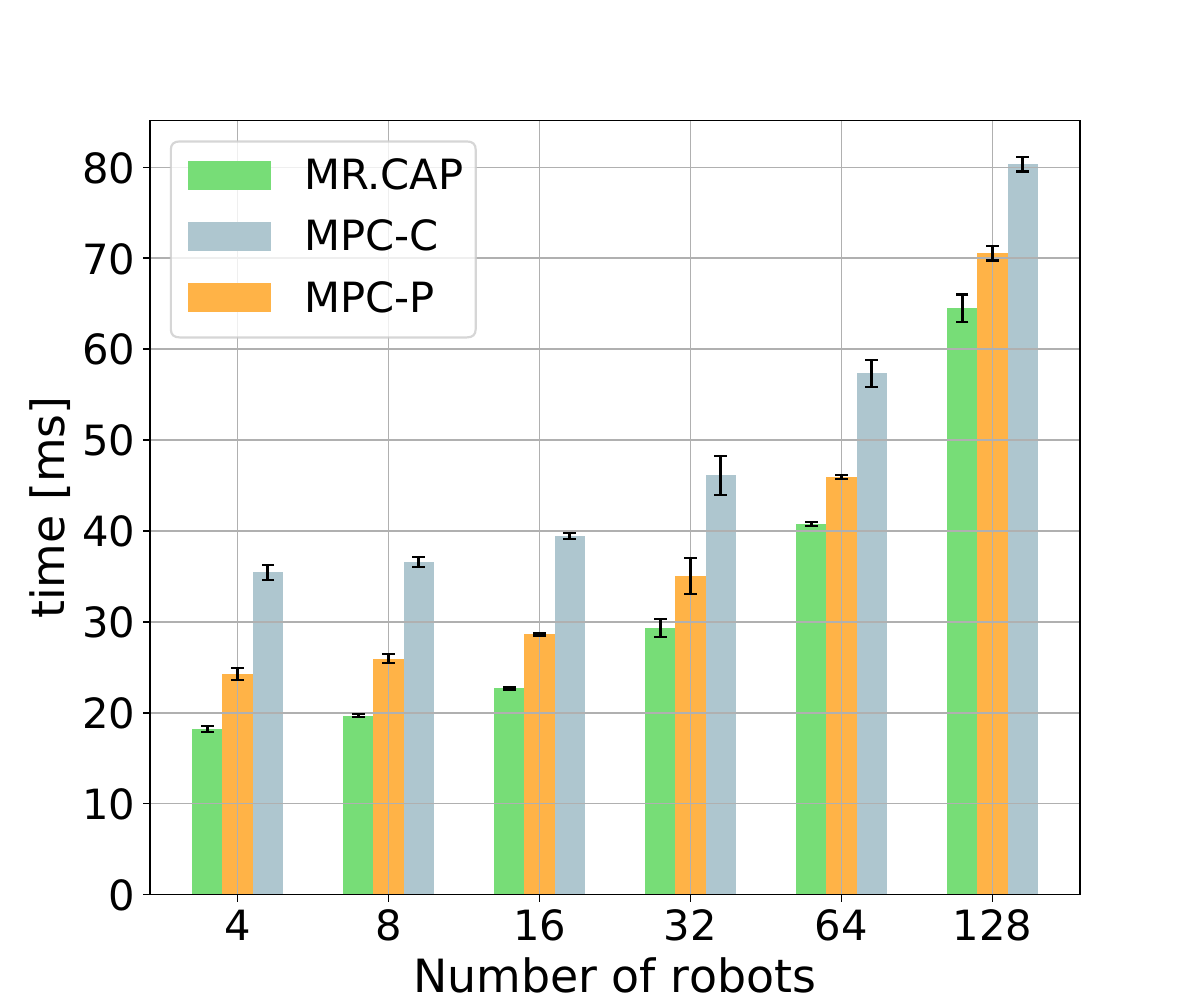}
        \label{fig:robot-scalability}
        \vspace{-.5 cm}
    \end{subfigure}
    \begin{subfigure}[b]{0.49\linewidth}
        \includegraphics[width=\linewidth]{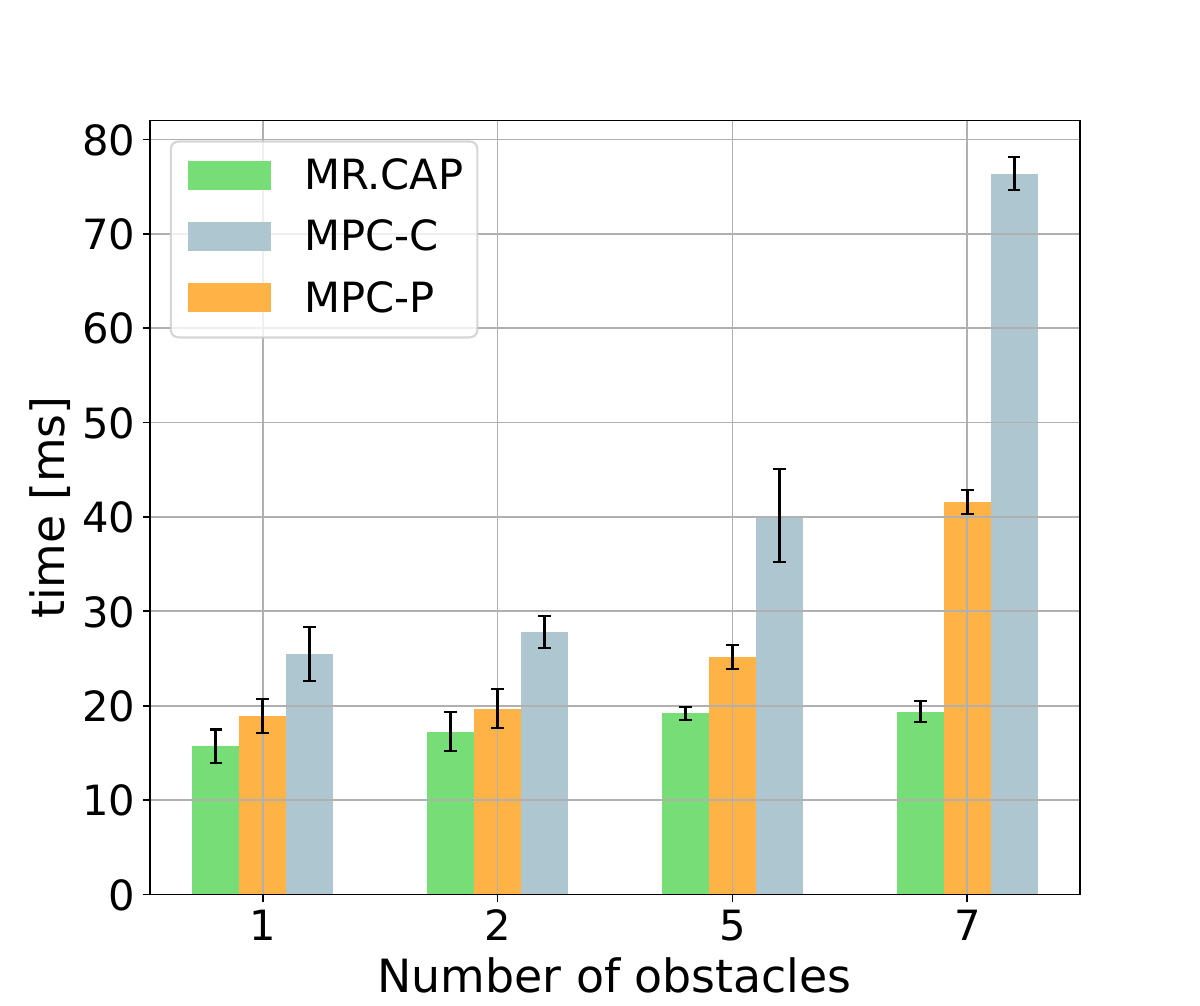}
        \label{fig:obstacle-scalability}
        \vspace{-.5 cm}
    \end{subfigure}
    \caption{Results of Experiment 1 showing the scalability with respect to the number of (left) robots, and (right) obstacles.} \label{fig:scal-exp}
\end{figure}

\begin{table}[]
\centering
\caption{Table outlining the benefits of using penalty (MPC-P) vs equalities (MPC-C) for the MPC approach. We selected the option that provided the best advantage to MPC. We also compare gradient (G) to gradient-free approaches (GF).}
\begin{tabular}{|c|c|c|c|c|}
\hline
           & \multicolumn{2}{c|}{MPC-C} & \multicolumn{2}{c|}{MPC-P} \\ \cline{2-5}
           & \multicolumn{1}{c|}{G} & {GF} & \multicolumn{1}{c|}{G} & {GF} \\ \hline
    Time (ms) & {34.93} & {89.28} & {23.95} & {55.584} \\ \hline
    Deviation (m) & {0.02} & {0.02} & {0.08} & {0.10} \\ \hline
    Distance to Goal (m) & {0.003} & {0.028} & {0.006} & {0.062} \\ \hline
    Path Length (m) & {9.04} & {8.99} & {9.37} & {9.15} \\ \hline
    Obstacle Free & {\ding{51}} & {\ding{55}} & {\ding{51}} & {\ding{55}} \\
\hline
\end{tabular}
\label{fig:equality}
\end{table}

\begin{table}[b]
\centering 
\caption{The best parameters for the algorithms by exhaustive search. The parameters serve as weights/covariances onto the associated errors in the cost function.} 
\resizebox{\columnwidth}{!}{%
\begin{tabular}{@{}c|ccc@{}}
\toprule
\textbf{Parameters} & {\textbf{Ours} (Covariance)} & {\textbf{MPC-C} (Weight)} & {\textbf{MPC-P} (Weight)}\\
\midrule
State               & ({1$m^2$}, {1$m^2$}, {0.2$rad^2$})$\times10^{-1}$ & 1$m^{-2}$ & 1$m^{-2}$\\
Terminal            & ({1$m^2$}, {1$m^2$}, {0.2$rad^2$})$\times10^{-1}$ & 1$\times10^{3}m^{-2}$ & 1$m^{-2}$\\
Control             & (1$m^2s^{-2}$, 1$m^2s^{-2}$)$\times10^{-1}$ & 1$\times10^{-3}m^{-2}s^2$ & 1$m^{-2}s^2$ \\
Motion Model        & (1$m^2$, 1$m^2$, $0.2rad^2$)$\times10^{-4}$ & -- & 1$\times10^{-1}m^{-2}$ \\
Obstacle            & 0.01             & --               & 1\\
Horizon             & --               & 2 steps         & 2 steps\\
Relative Tolerance  & 1$\times10^{-2}$ & 1$\times10^{-4}$ & 1$\times10^{-2}$\\
Absolute Tolerance  & 1$\times10^{-2}$ & 1$\times10^{-4}$ & 1$\times10^{-2}$\\
Error Tolerance     & 1$\times10^{-2}$ & 1$\times10^{-4}$ & 1$\times10^{-2}$\\
\bottomrule
\end{tabular}
}
\label{fig: parameters}
\end{table}
\begin{table}[!]
\centering
\caption{Simulation Results: Payload navigating from (0,0) to (5,3) in Gazebo environment. {The length of the initial path estimate} is 5.831m. {MPC-C (constraint), MPC-P (penalty), and hierarchal quadratic programming HQP~\cite{hal-geometric-approach} are shown.}}
    \resizebox{0.8\linewidth}{!}{%
\begin{tabular}{@{}l|cccc@{}}
\toprule
                             & \textbf{Ours}     & \textbf{MPC-C}   & {\textbf{MPC-P}}        & \textbf{HQP~\cite{hal-geometric-approach}}            \\ \midrule
Avg. Deviation (m)           & \textbf{0.015} &  0.017            & {0.019}                    & -                      \\
Max Inter-Robot Error (m)    & \textbf{0.016}    & 0.05            & {0.05}                       & 0.017                  \\
Path Length (m)              & \textbf{{5.824}}    & {5.825}           & {5.825}                       & -                      \\ 
Dist. to Goal (m)            & \textbf{0.017}    & 0.026            &  {0.023}               & - \\
\bottomrule
\end{tabular}
}
\label{fig: gazebo}
\end{table}

Robots traversed from (-2, 0) to (7, 0) with obstacles at (1, 0.4), (2, -0.4), (3, -0.6), (4, 0.7) (5, 0.4), all in meters.  We measured optimization time, average deviation from the {initial} path, and the path length. Results are plotted in Fig.~\ref{fig:scal-exp}. The MPC results in the figure are obtained from MPC-C and MPC-P with gradient-based solvers, as they have been proven to be faster and more reliable (see Table~\ref{fig:equality}).

As shown in Fig.~\ref{fig:scal-exp}, under the reduced formulation of the problem, the factor graph representation consistently outperforms MPC-C and MPC-P in all experiments. The optimization time of the three algorithms all scaled linearly in the robot scalability experiment, Fig.~\ref{fig:scal-exp}-(left). Such increases in times are associated with some overhead computing individual robot controls that increase with the number of robots which are minimal and can be further optimized.

It can also be seen from Fig.~\ref{fig:scal-exp}-(right) that the factor graph representation has allowed the optimization to scale better as more obstacles are added to the environment. We notice the optimization time being 38.4\%, 37.9\%, 52.1\% and \textcolor{black}{76.1\%} shorter than those of MPC-C, and 17.1\%, 12.2\%, 23.7\% and \textcolor{black}{56.1\%} shorter than those of MPC-P in this obstacle scalability experiment. Our approach consistently reaches the goal, while MPC-C's distances to the goal are 0.003$m$, 0.001$m$, 0.002$m$, and \textcolor{black}{0.002$m$}, and MPC-P's 0.002$m$, 0.006$m$, 0.006$m$, and \textcolor{black}{0.01$m$}. One downside to our approach is that it yields path lengths longer than MPC-C's by 2.99\%, 3.72\%, and 7.16\%, and \textcolor{black}{6.56\%} and MPC-P's by 1.73\%, 0.81\%, 3.38\%, and \textcolor{black}{3.06\%}. Increasing obstacles from 1 to 2, the optimization times for our approach, MPC-C, and MPC-P increased by 10.1\%, 9.2\%, and 3.95\%. Increasing from 2 to \textcolor{black}{7}, the optimization times increased by another \textcolor{black}{5.84\%, 174.4\%, and 111.5\%}

\begin{figure*}[h]
    \centering
    \begin{subfigure}{0.3\textwidth}
        \centering
        \includegraphics[width=\textwidth]{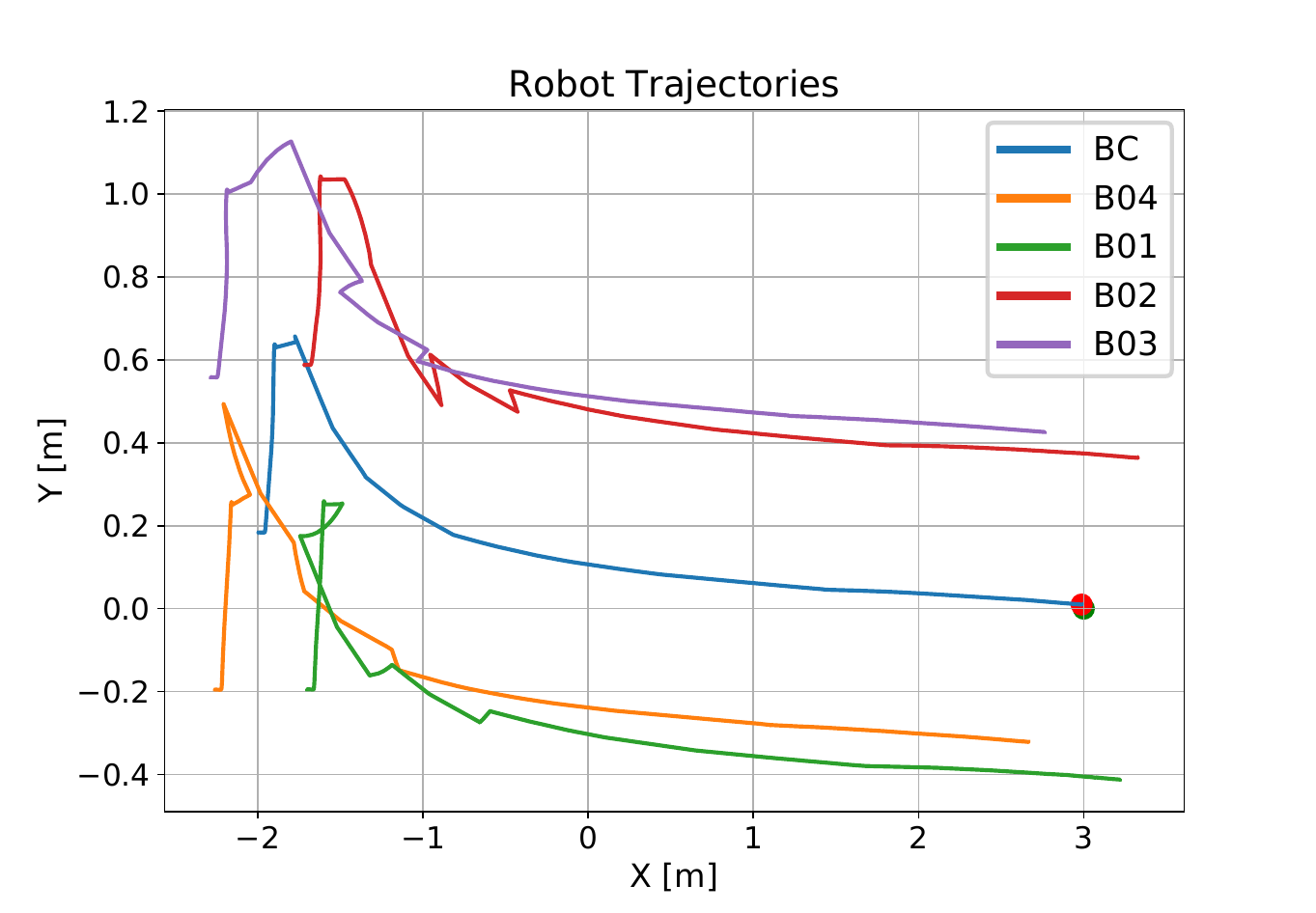} 
        \includegraphics[width=\textwidth]{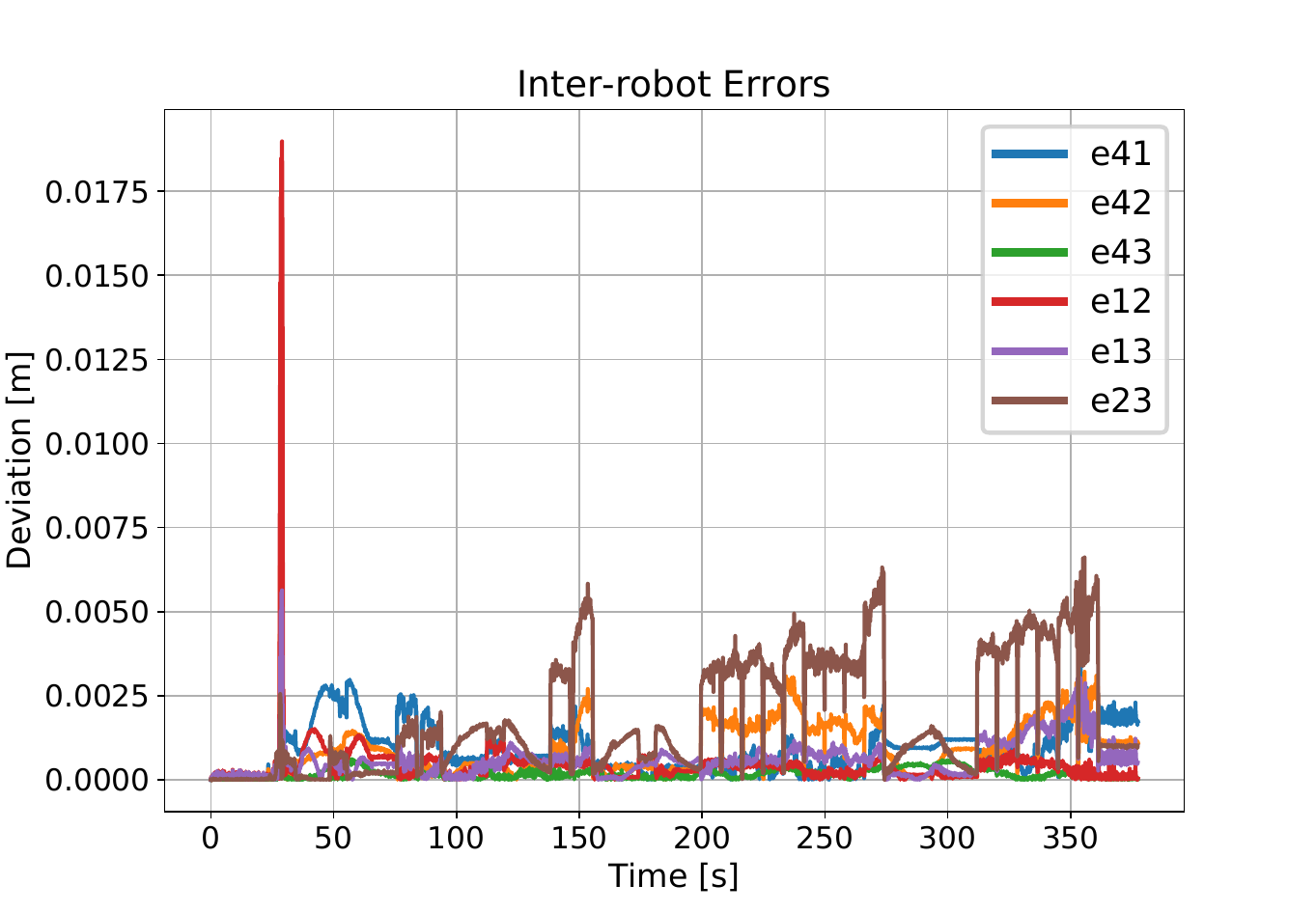}
        \label{fig: distur}
    \end{subfigure}%
    \begin{subfigure}{0.3\textwidth}
        \centering
        \includegraphics[width=\textwidth]{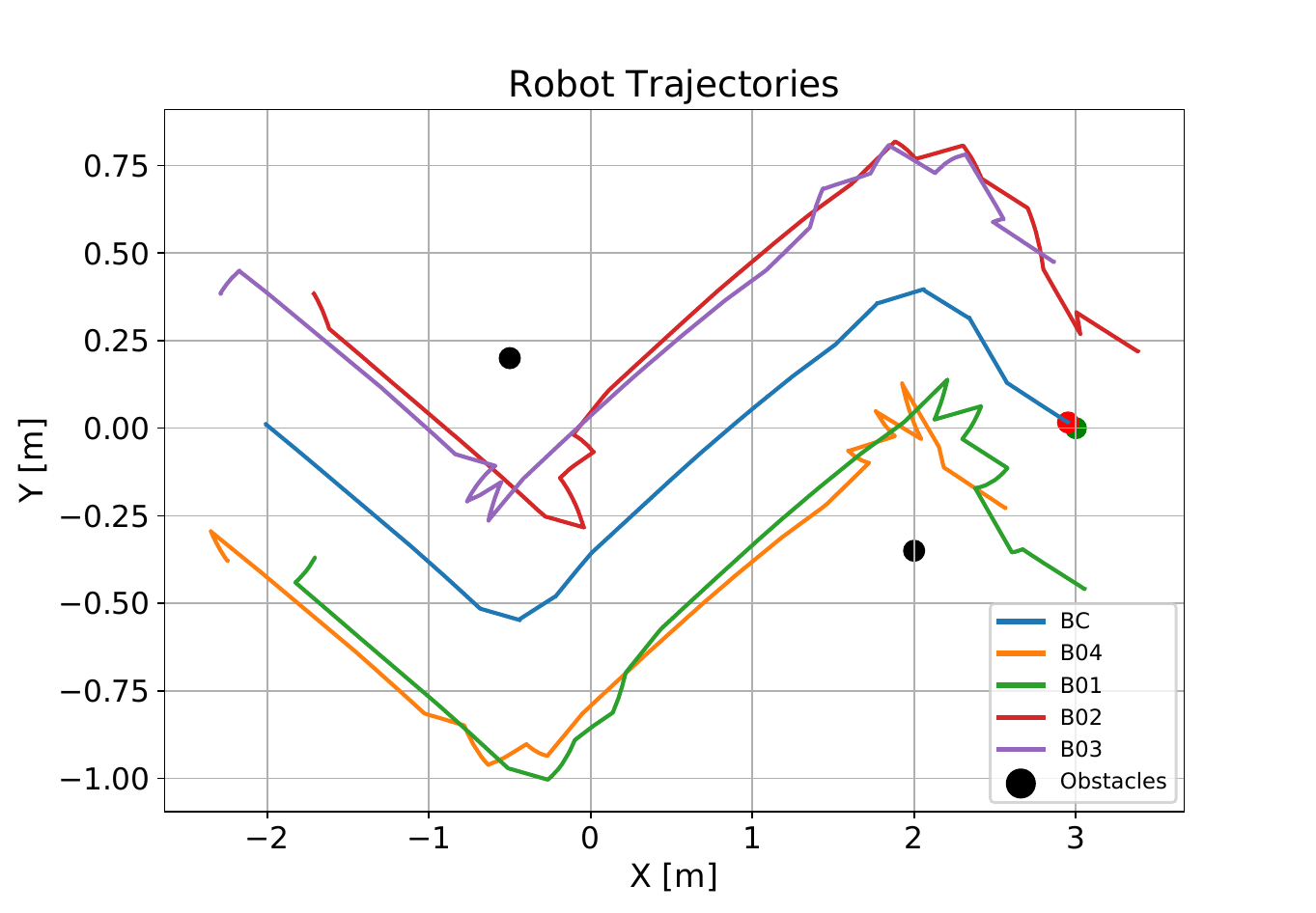}
        \includegraphics[width=\textwidth]{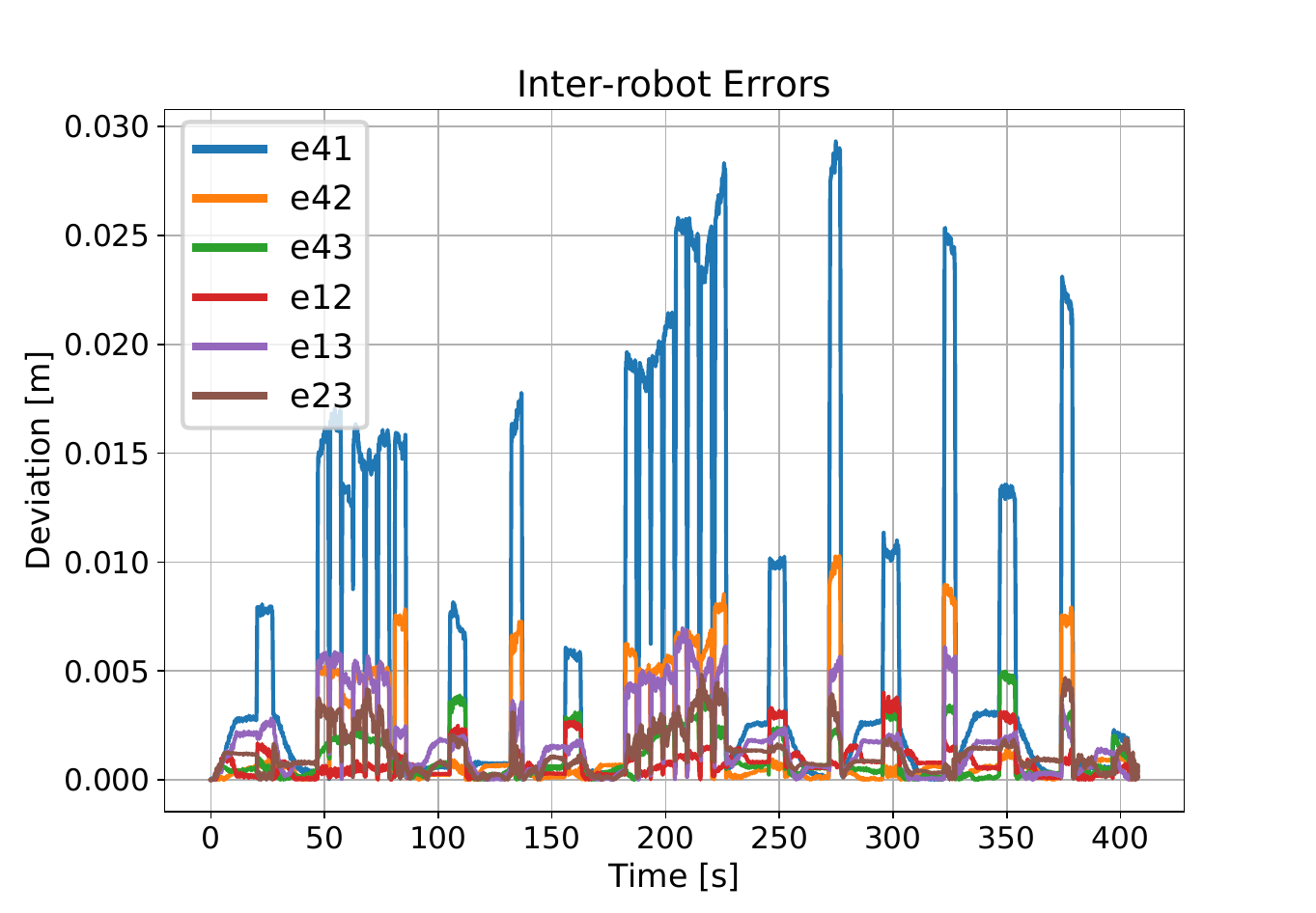}
        \label{fig: obs aoid}
    \end{subfigure}%
    \begin{subfigure}{0.3\textwidth}
        \centering
        \includegraphics[width=\textwidth]{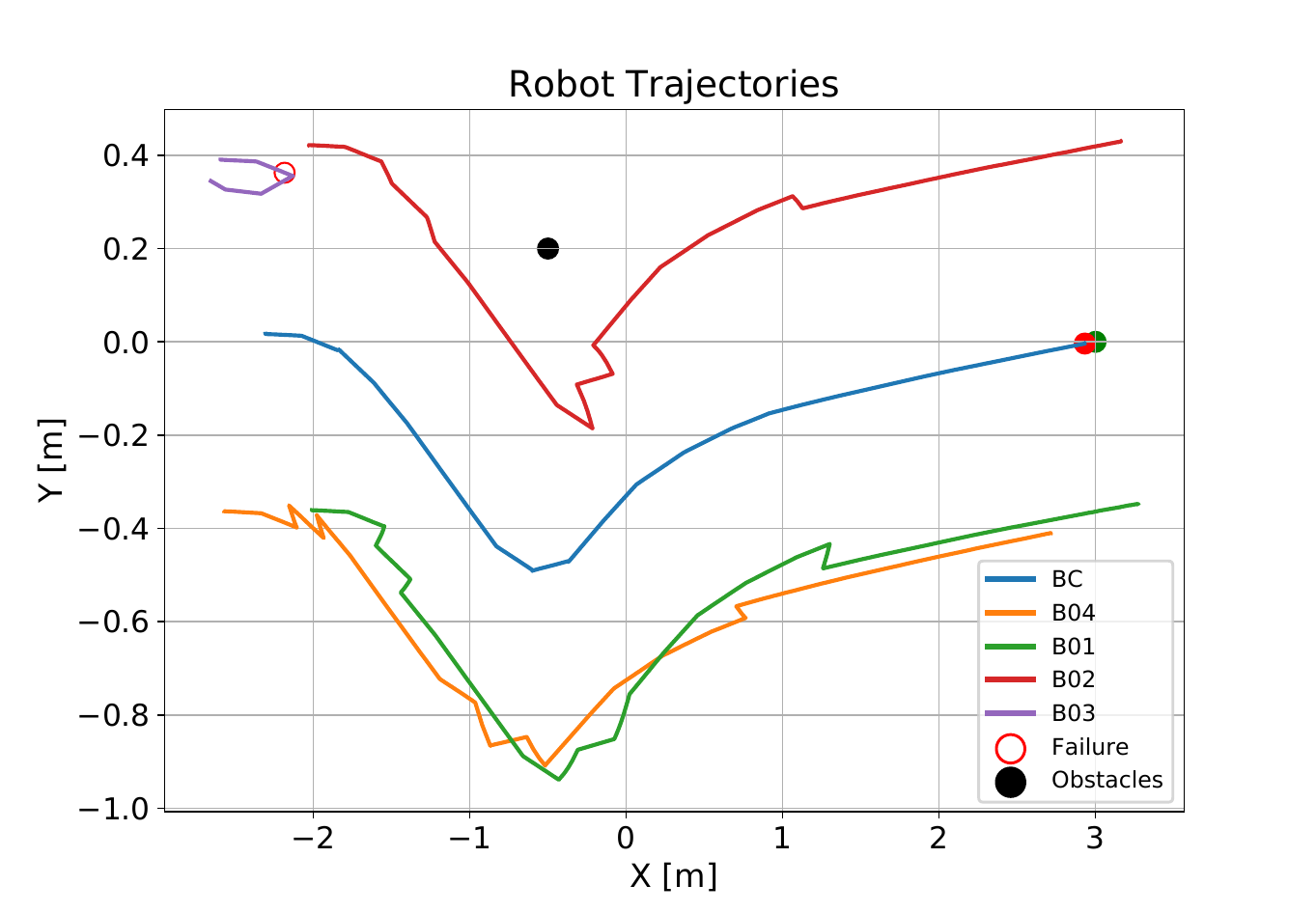}
        \includegraphics[width=\textwidth]{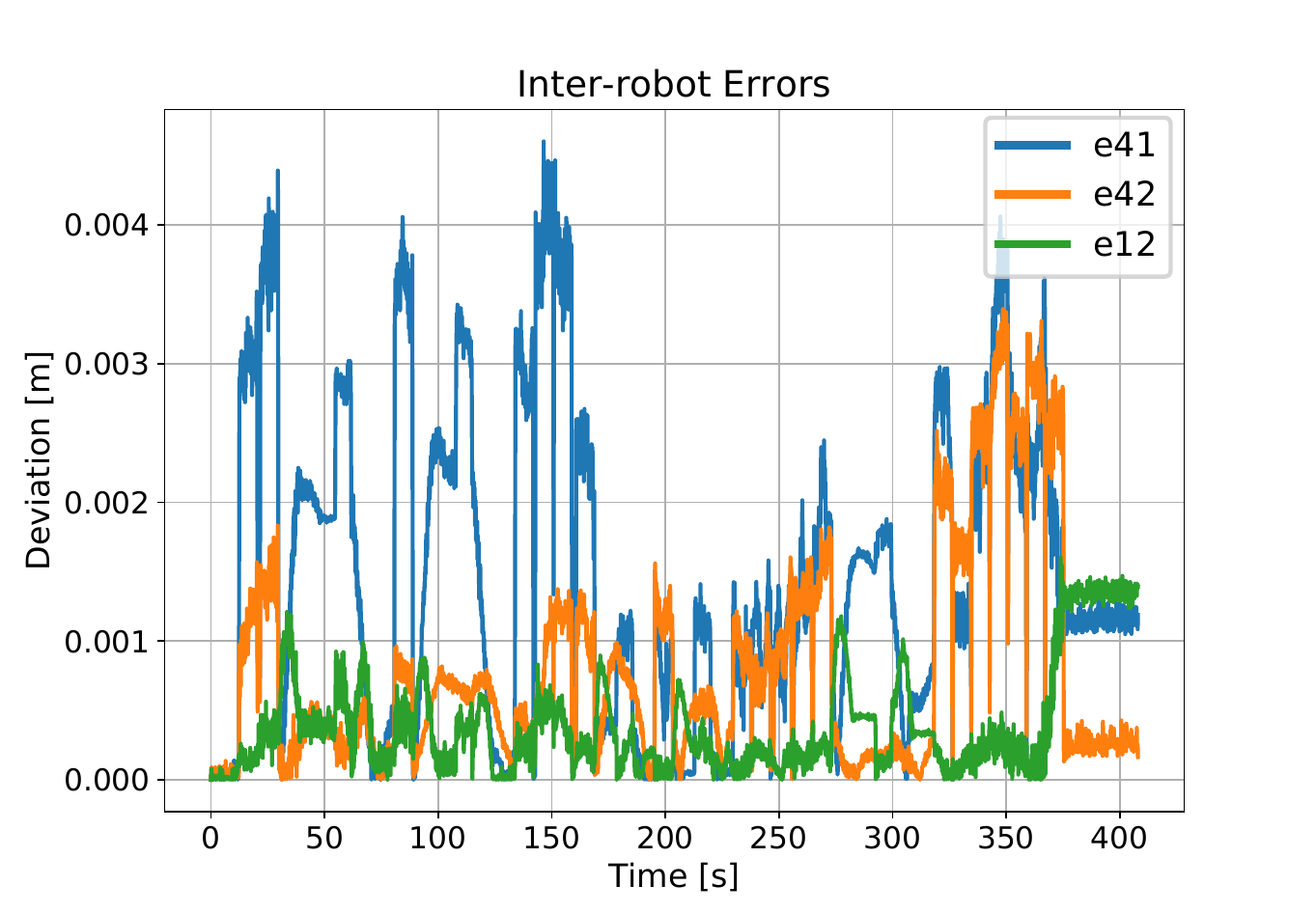}
        \label{fig: failure}
    \end{subfigure}
    \vspace{-0.4 cm}
    \caption{Hardware trial results. A payload was pushed by four Turtlebot3 Waffle Pi robots from random start points near the origin (0, 0) to the same goal (3, 0). The three columns correspond to the three types of tasks: (left column) reaching the goal while undergoing large disturbances, (middle column) obstacle avoidance, and (right column) robot failure during obstacle avoidance. The system is consistently able to recover and reach the goal within 1cm accuracy.}
        \label{fig: hardwaredem}
\end{figure*}

\subsection{Experiment 2: Gazebo Simulation} 
In addition to MPC, we compare with another baseline, hierarchical quadratic programming (HQP)~\cite{hal-geometric-approach}. {Since the code for HPQ is not open-source,} we used the same Gazebo~\cite{gazebo} world with identical start/goal positions and geometric parameters described in~\cite{hal-geometric-approach} {to run our algorithm.} We compare inter-robot errors, as deviation and temporal results were not provided {in~\cite{hal-geometric-approach}}. As shown in Table~\ref{fig: gazebo}, our approach outperforms HQP in terms of inter-robot errors, and MPC} in average path deviation, inter-robot error, and distance to goal.

\subsection{Experiment 3: Real-world Robots}
As {shown} in Fig.~\ref{fig:two-robots}, four TurtleBot3 Waffle Pi robots are deployed with {an object} lifted on top of them. These robots are connected through revolute joints {to the object}, as they allow for fixed geometry of the system while also maintaining the system's ability to fully rotate. 
ROS2 is used as the communication middleware. We obtain robot poses from a Vicon motion capture system with 1mm accuracy. We present {three} scenarios to evaluate the performance and demonstrate the robustness of {the} framework: i) disturbances, ii) obstacle avoidance, and iii) robot failure. The results are presented in Fig.~\ref{fig: hardwaredem}. {The rows show robot paths and the inter-robot errors.} {Fig.~\ref{fig: hardwaredem}-(left column) shows the case where} the system was disturbed by 40cm, {at time 30 sec}, while executing a path and was able to recover from the disturbance within a few steps to reach the goal with sub-1 cm accuracy. 
{Fig.~\ref{fig: hardwaredem}-(middle column) shows} the system tasked with obstacle avoidance while on a mission, and successfully avoided the obstacles reaching the goal point within 1cm accuracy. 
{Fig.~\ref{fig: hardwaredem}-(right column) shows a robot failure.} One robot was disabled and removed from the trial. The system was able to compensate and reach the goal within 1cm accuracy.

\section{{Conclusions}}  \label{sec:conclusions}

In this paper, we proposed a novel technique for multi-robot joint planning and control, leveraging the scalability and robustness of factor graphs and the inherent forward-looking nature of the planning and control problems. We make use of a single optimization to reduce optimization times, which is further aided by reducing the dimensionality to yield a highly scalable approach. We compare our approach against state-of-the-art and consistently outperformed previous approaches.  Furthermore, our approach is highly extensible. Adding new constraints on the system is a matter of simply adding factors to the graph, and dynamics can be modified through simple modification of the motion model while the optimization remains the same. Comparative results show an improvement in optimization time, observing a reduction of 50.1\% {in scalability experiments}. Our algorithm consistently reduced the average path deviation and optimization time and produced a shorter distance to the goal. Similarly, our hardware experiments {showed} the robustness of our approach, with demonstrations of disturbances, obstacle avoidance, and robot failures. Scalability experiments {showed that the} computation time does not significantly increase with increased environment complexity. This is due to both the formulation and the use of factor graphs; as MPC approaches made use of similar formulation and observed similar scaling, {while the factor graph allowed us to outperform MPC}.  

In future works, we will integrate Gaussian processes into the control problem to allow for smoother paths and a continuous-time representation. Another direction is extending the work to motions with higher degrees of freedom.

\bibliographystyle{ieeetr}
\bibliography{main}

\begin{thebibliography}{10}

\bibitem{Parker2009}
L.~E. Parker, {\em Multiple Mobile Robot Teams, Path Planning and Motion Coordination}, pp.~5783--5800.
\newblock Springer, 2009.

\bibitem{STEAP}
M.~Mukadam, J.~Dong, F.~Dellaert, and B.~Boots, ``Steap: simultaneous trajectory estimation and planning,'' {\em Autonomous Robots}, vol.~43, no.~2, pp.~415--434, 2018.

\bibitem{halsted2021survey}
T.~Halsted, O.~Shorinwa, J.~Yu, and M.~Schwager, ``A survey of distributed optimization methods for multi-robot systems,'' {\em ArXiv}, vol.~abs/2103.12840, 2021.

\bibitem{mclurkin2015}
G.~Habibi, Z.~Kingston, W.~Xie, M.~Jellins, and J.~McLurkin, ``Distributed centroid estimation and motion controllers for collective transport by multi-robot systems,'' in {\em {ICRA}}, pp.~1282--1288, 2015.

\bibitem{hal-geometric-approach}
D.~Koung, O.~Kermorgant, I.~Fantoni, and L.~Belouaer, ``Cooperative multi-robot object transportation system based on hierarchical quadratic programming,'' {\em IEEE RA-L}, vol.~6, no.~4, pp.~6466--6472, 2021.

\bibitem{factor-graphs-for-perception}
F.~Dellaert and M.~Kaess, {\em Factor Graphs for Robot Perception}.
\newblock Foundations and Trends in Robotics Series, Now Publishers, 2017.

\bibitem{AUGLAG}
A.~R. Conn, N.~I.~M. Gould, and P.~Toint, ``A globally convergent augmented {Lagrangian} algorithm for optimization with general constraints and simple bounds,'' {\em {SIAM} Journal on Numerical Analysis}, vol.~28, pp.~545--572, 1991.

\bibitem{Powell1994ADS}
M.~J.~D. Powell, {\em A Direct Search Optimization Method That Models the Objective and Constraint Functions by Linear Interpolation}, pp.~51--67.
\newblock Dordrecht: Springer Netherlands, 1994.

\bibitem{astar_og}
P.~E. Hart, N.~J. Nilsson, and B.~Raphael, ``A formal basis for the heuristic determination of minimum cost paths,'' {\em IEEE SMC}, vol.~4, no.~2, pp.~100--107, 1968.

\bibitem{choset2005principles}
H.~M. Choset, S.~Hutchinson, K.~M. Lynch, G.~Kantor, W.~Burgard, L.~E. Kavraki, and S.~Thrun, {\em {Principles of Robot Motion: theory, algorithms, and implementation}}.
\newblock MIT press, 2005.

\bibitem{rrt}
S.~Lavalle and J.~Kuffner, ``Rapidly-exploring random trees: Progress and prospects,'' {\em Algorithmic and Computational Robotics}, 2000.

\bibitem{kavraki1994probabilistic}
L.~E. Kavralu, P.~Svestka, J.-C. Latombe, and M.~H. Overmars, ``Probabilistic roadmaps for path planning in high-dimensional configuration spaces,'' {\em IEEE T-RO}, 1996.

\bibitem{chomp}
M.~Zucker, N.~Ratliff, A.~D. Dragan, M.~Pivtoraiko, M.~Klingensmith, C.~M. Dellin, J.~A. Bagnell, and S.~S. Srinivasa, ``Chomp: Covariant hamiltonian optimization for motion planning,'' {\em IJRR}, vol.~32, no.~9-10, pp.~1164--1193, 2013.

\bibitem{T-CHOMP}
A.~Byravan, B.~Boots, S.~Srinivasa, and D.~Fox, ``Space-time functional gradient optimization for motion planning,'' in {\em ICRA}, pp.~6499 -- 6506, 05 2014.

\bibitem{multi-chomp}
K.~He, E.~Martin, and M.~Zucker, ``Multigrid chomp with local smoothing,'' in {\em 2013 13th IEEE-RAS (Humanoids)}, pp.~315--322, 2013.

\bibitem{STOMP}
M.~Kalakrishnan, S.~Chitta, E.~Theodorou, P.~Pastor, and S.~Schaal, ``Stomp: Stochastic trajectory optimization for motion planning,'' in {\em 2011 IEEE ICRA}, pp.~4569--4574, 2011.

\bibitem{toussaint2009}
M.~Toussaint, ``Robot trajectory optimization using approximate inference,'' in {\em ICML}, p.~1049–1056, 2009.

\bibitem{gpmp2}
J.~Dong, M.~Mukadam, F.~Dellaert, and B.~Boots, ``Motion planning as probabilistic inference using gaussian processes and factor graphs.,'' in {\em RSS}, vol.~12, 2016.

\bibitem{TrajOpt2}
J.~Schulman, Y.~Duan, J.~Ho, A.~Lee, I.~Awwal, H.~Bradlow, J.~Pan, S.~Patil, K.~Goldberg, and P.~Abbeel, ``Motion planning with sequential convex optimization and convex collision checking,'' {\em IJRR}, vol.~33, no.~9, pp.~1251--1270, 2014.

\bibitem{gpmp}
M.~Mukadam, X.~Yan, and B.~Boots, ``Gaussian process motion planning,'' in {\em 2016 IEEE (ICRA)}, pp.~9--15, 2016.

\bibitem{bazzana_handling_2022}
B.~Bazzana, T.~Guadagnino, and G.~Grisetti, ``Handling {Constrained} {Optimization} in {Factor} {Graphs} for {Autonomous} {Navigation},'' 2022.

\bibitem{DFG}
M.~Xie, A.~Escontrela, and F.~Dellaert, ``A factor-graph approach for optimization problems with dynamics constraints,'' {\em arXiv preprint arXiv:2011.06194}, 2020.

\bibitem{ITOMP}
C.~Park, J.~Pan, and D.~Manocha, ``{ITOMP}: Incremental trajectory optimization for real-time replanning,'' in {\em ICAPS}, 2012.

\bibitem{patwardhan2023distributed}
A.~Patwardhan and A.~J. Davison, ``A distributed multi-robot framework for exploration, information acquisition and consensus,'' {\em arXiv preprint arXiv:2310.01930}, 2023.

\bibitem{murai2022robotweb}
R.~Murai, J.~Ortiz, S.~Saeedi, P.~H.~J. Kelly, and A.~J. Davison, ``{A Robot Web for Distributed Many-Device Localisation},'' {\em IEEE T-RO}, vol.~40, pp.~121--138, 2024.

\bibitem{spasojevic_active_nodate}
I.~Spasojevic, X.~Liu, A.~Ribeiro, G.~J. Pappas, and V.~Kumar, ``Active {Collaborative} {Localization} in {Heterogeneous} {Robot} {Teams},'' {\em arXiv preprint arXiv:2305.18193}, 2023.

\bibitem{guerrero-bonilla_formations_2017}
L.~Guerrero-Bonilla, A.~Prorok, and V.~Kumar, ``Formations for {Resilient} {Robot} {Teams},'' {\em IEEE RA-L}, vol.~2, no.~2, pp.~841--848, 2017.

\bibitem{gnn}
Q.~Li, F.~Gama, A.~Ribeiro, and A.~Prorok, ``Graph neural networks for decentralized multi-robot path planning,'' in {\em IROS}, pp.~11785--11792, 2020.

\bibitem{cao_path_nodate}
M.~Cao, K.~Cao, S.~Yuan, K.~Liu, Y.~L. Wong, and L.~Xie, ``Path {Planning} for {Multiple} {Tethered} {Robots} {Using} {Topological} {Braids},'' {\em RSS}, 2023.

\bibitem{gbp-planner}
A.~Patwardhan, R.~Murai, and A.~J. Davison, ``Distributing collaborative multi-robot planning with gaussian belief propagation,'' {\em IEEE Robotics and Automation Letters}, vol.~8, no.~2, pp.~552--559, 2023.

\bibitem{pierson_cooperative_2016}
A.~Pierson, A.~Ataei, I.~C. Paschalidis, and M.~Schwager, ``Cooperative multi-quadrotor pursuit of an evader in an environment with no-fly zones,'' in {\em {ICRA}}, pp.~320--326, May 2016.

\bibitem{dellaert-fg-approach}
D.-N. Ta, M.~Kobilarov, and F.~Dellaert, ``A factor graph approach to estimation and model predictive control on unmanned aerial vehicles,'' in {\em 2014 ICUAS}, pp.~181--188, 2014.

\bibitem{saravanos_distributed_nodate}
A.~D. Saravanos, Y.~Li, and E.~A. Theodorou, ``Distributed {Hierarchical} {Distribution} {Control} for {Very}-{Large}-{Scale} {Clustered} {Multi}-{Agent} {Systems},'' {\em RSS}, 2023.

\bibitem{yan_decentralized_2021}
L.~Yan, T.~Stouraitis, and S.~Vijayakumar, ``Decentralized {Ability}-{Aware} {Adaptive} {Control} for {Multi}-{Robot} {Collaborative} {Manipulation},'' {\em IEEE RAL}, vol.~6, no.~2, pp.~2311--2318, 2021.

\bibitem{HE20209859}
Y.~He, M.~Wu, and S.~Liu, ``An optimisation-based distributed cooperative control for multi-robot manipulation with obstacle avoidance,'' {\em IFAC-PapersOnLine}, vol.~53, no.~2, pp.~9859--9864, 2020.

\bibitem{ogasawara1996}
Y.-H. Liu, S.~Arimoto, and T.~Ogasawara, ``Decentralized cooperation control: Non-communication object handling,'' in {\em {ICRA}}, vol.~3, pp.~2414--2419 vol.3, 1996.

\bibitem{jennings1995}
D.~Rus, B.~Donald, and J.~Jennings, ``Moving furniture with teams of autonomous robots,'' in {\em {IROS}}, vol.~1, pp.~235--242 vol.1, 1995.

\bibitem{lujak_distributed_2010}
M.~Lujak, ``A {Distributed} {Coordination} {Model} for {Multi}-{Robot} {Box} {Pushing},'' {\em IFAC Proceedings Volumes}, vol.~43, no.~4, pp.~120--125, 2010.

\bibitem{daniela2015}
J.~Alonso-Mora, R.~Knepper, R.~Siegwart, and D.~Rus, ``Local motion planning for collaborative multi-robot manipulation of deformable objects,'' in {\em {ICRA}}, pp.~5495--5502, 2015.

\bibitem{schwager2015}
Z.~Wang and M.~Schwager, ``Multi-robot manipulation with no communication using only local measurements,'' in {\em {CDC}}, pp.~380--385, 2015.

\bibitem{schwager2016}
Z.~Wang and M.~Schwager, ``Kinematic multi-robot manipulation with no communication using force feedback,'' in {\em ICRA}, pp.~427--432, 2016.

\bibitem{schwagerants2016}
Z.~Wang and M.~Schwager, ``Force-amplifying n-robot transport system (force-ants) for cooperative planar manipulation without communication,'' {\em IJRR}, vol.~35, no.~13, pp.~1564--1586, 2016.

\bibitem{culbertson_decentralized_2018}
P.~Culbertson and M.~Schwager, ``Decentralized {Adaptive} {Control} for {Collaborative} {Manipulation},'' in {\em {ICRA}}, pp.~278--285, 2018.

\bibitem{data-distributed}
R.~T. Fawcett, L.~Amanzadeh, J.~Kim, A.~D. Ames, and K.~A. Hamed, ``Distributed data-driven predictive control for multi-agent collaborative legged locomotion,'' in {\em ICRA}, pp.~9924--9930, 2023.

\bibitem{data-driven-ral}
R.~T. Fawcett, K.~Afsari, A.~D. Ames, and K.~A. Hamed, ``Toward a data-driven template model for quadrupedal locomotion,'' {\em IEEE RA-L}, vol.~7, no.~3, pp.~7636--7643, 2022.

\bibitem{schwager2020-control-planning}
O.~Shorinwa and M.~Schwager, ``Scalable collaborative manipulation with distributed trajectory planning,'' in {\em IROS}, pp.~9108--9115, 2020.

\bibitem{schwager2021trajopt}
O.~Shorinwa and M.~Schwager, ``Distributed contact-implicit trajectory optimization for collaborative manipulation,'' in {\em MRS}, pp.~56--65, 2021.

\bibitem{gtsam}
{Frank Dellaert and {GTSAM} Contributors}, ``{GTSAM}.'' \url{https://github.com/borglab/gtsam}, May 2022.

\bibitem{LM_OG_PAPER}
D.~W. Marquardt, ``An algorithm for least-squares estimation of nonlinear parameters,'' {\em SIAM}, vol.~11, no.~2, pp.~431--441, 1963.

\bibitem{Yang_2021}
S.~Yang, G.~Chen, Y.~Zhang, H.~Choset, and F.~Dellaert, ``Equality constrained linear optimal control with factor graphs,'' in {\em ICRA}, IEEE, May 2021.

\bibitem{barrierMPC}
R.~Grandia, F.~Farshidian, R.~Ranftl, and M.~Hutter, ``Feedback {MPC} for torque-controlled legged robots,'' in {\em IROS}, pp.~4730--4737, 2019.

\bibitem{Xie2020AFA}
M.~Xie, A.~Escontrela, and F.~Dellaert, ``A factor-graph approach for optimization problems with dynamics constraints,'' {\em ArXiv}, vol.~abs/2011.06194, 2020.

\bibitem{bergou_convergence_2020}
E.~H. Bergou, Y.~Diouane, and V.~Kungurtsev, ``Convergence and {Complexity} {Analysis} of a {Levenberg}–{Marquardt} {Algorithm} for {Inverse} {Problems},'' {\em Journal of Optimization Theory and Applications}, vol.~185, pp.~927--944, June 2020.

\bibitem{auglag_complexity}
G.~N. Grapiglia and Y.-x. Yuan, ``{On the complexity of an augmented Lagrangian method for nonconvex optimization},'' {\em IMA Journal of Numerical Analysis}, vol.~41, pp.~1546--1568, 07 2020.

\bibitem{gazebo}
{{Open Robotics}}, ``Gazebo: Simulate before you build.'' \url{{https://gazebosim.org/home}}, 2023.

\bibitem{NLopt}
S.~G. Johnson, ``The {NLopt} nonlinear-optimization package.'' \url{{http://github.com/stevengj/nlopt}}, 2007.

\end{thebibliography}

\end{document}